\documentclass[10pt,twocolumn,letterpaper]{article}

\usepackage{wacv}
\usepackage{times}
\usepackage{epsfig}
\usepackage{graphicx}
\usepackage{amsmath}
\usepackage{amssymb}
\usepackage{xcolor}

\usepackage[utf8]{inputenc} 
\usepackage[T1]{fontenc}    
\usepackage{url}            
\usepackage{booktabs}       
\usepackage{amsfonts}       
\usepackage{nicefrac}       
\usepackage{microtype}      
\usepackage{enumerate}
\usepackage{algpseudocode,algorithm,algorithmicx}  

\usepackage[colorlinks=true,linkcolor=blue,bookmarks=false]{hyperref}

\usepackage{color,soul}
\usepackage[nomargin,inline,marginclue,draft]{fixme}

\usepackage{amsmath,amssymb}

\DeclareMathOperator{\loss}{\mathcal{L}}
\DeclareMathOperator{\E}{\mathbb{E}}

\algrenewcommand\algorithmicindent{2mm}%

\def\wacvPaperID{207} 

\wacvfinalcopy

\ifwacvfinal\fi
\begin{document}

\title{Learning Multimodal Representations for Unseen Activities}

\author{AJ Piergiovanni$^1$ \hspace{2cm} Michael S. Ryoo$^{1,2}$ \\
$^1$Indiana University, $^2$Stony Brook University\\
{\tt\small \{ajpiergi,mryoo\}@indiana.edu}
}

\maketitle
\ifwacvfinal\thispagestyle{empty}\fi

\begin{abstract}
 We present a method to learn a joint multimodal representation space that enables recognition of unseen activities in videos. We first compare the effect of placing various constraints on the embedding space using paired text and video data. We also propose a method to improve the joint embedding space using an adversarial formulation, allowing it to benefit from \emph{unpaired} text and video data.  By using unpaired text data, we show the ability to learn a representation that better captures unseen activities.
 In addition to testing on publicly available datasets, we introduce a new, large-scale text/video dataset.
 We experimentally confirm that using paired and unpaired data to learn a shared embedding space benefits three difficult tasks (i) zero-shot activity classification, (ii) unsupervised activity discovery, and (iii) unseen activity captioning, outperforming the state-of-the-arts.
\end{abstract}

\section{Introduction}
Videos contain multiple data sources, such as visual, audio and text/caption data. Each data modality has distinct statistical properties capturing different aspects of the event. Current state-of-the-art activity recognition models \cite{carreira2017quo, tran2017closer} only take visual data and class labels as input, which limits the information the model can learn from. 
For example, the sentence `a group of men play basketball outdoors' contains rich information, such as `outdoors' and `group of men' compared to just the activity class label of `basketball.' We desire to use such additional information to learn better representations and by doing so, we show that the learned representations are able to generalize to unseen activities (i.e., zero-shot learning).



\begin{figure}
    \centering
    \includegraphics[width=\linewidth]{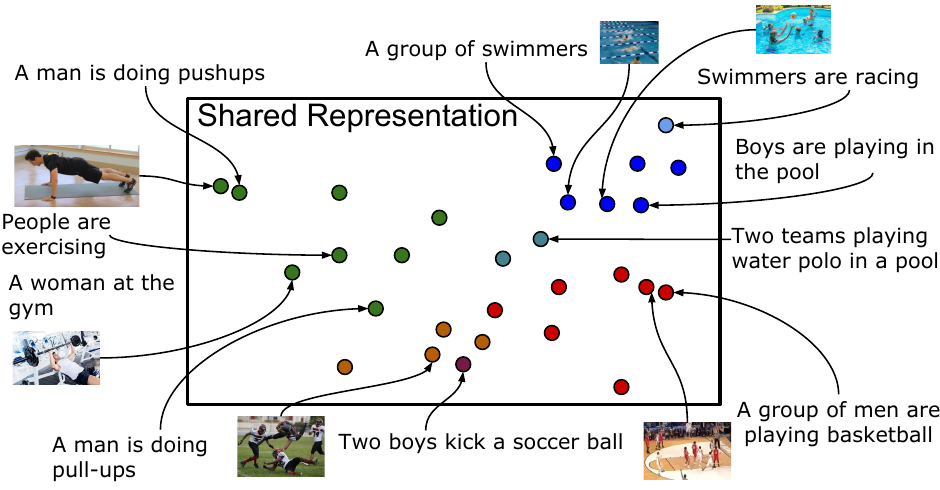}
    \caption{Taking advantage of both text and video data allows for learning of a shared representation. By utilizing unpaired text and video data, the representation naturally captures the relationships between different activities, based on the underlying relationships in word embeddings and video representations. The colors represent different activity classes of the video or sentence (e.g., various sports, pool activities, and exercises).}
    \label{fig:concept}
\end{figure}

We explore multimodal learning from video and language data, each starting with its own representation. Video data is represented as a sequence of images (spatio-temporal pixel data) while text is represented as a sequence of word embeddings (temporal data). Learning a shared representation allows for modeling the highly non-linear relationships between these modalities, capturing structure present in both video and textual data. Further, using a shared representation enables capturing similarities between concepts (e.g., basketball and volleyball both being sports with a ball) within its space by relying on either modality, even when the data is unpaired. This allows the representation to benefit from concepts not seen in both modalities during training. For example, we show taking advantage of relationships between words in pre-trained word embeddings \cite{mikolov2013efficient} help recognize activities with no video examples.
By learning a shared representation space, we transfer such relationships to video representations of potentially unseen activities. An conceptual overview of the approach is shown in Fig. \ref{fig:concept}.


Many existing approaches to both zero-shot and embedding space learning require paired data examples (e.g., examples and labeled attributes), which can be expensive to obtain. By taking advantage of adversarial learning \cite{goodfellow2014generative}, we are able effectively augment our method with \textbf{unpaired} data (i.e., random sentences and random videos without any labels or correspondence) to further improve our learned representation. By introducing many random videos and text data, we show that we are able learn representations that better capture unseen activities, without requiring any further annotations.

In this paper, we design a method capable of learning joint video/language representation using both paired and unpaired data. We experimentally confirm its benefit to three challenging tasks: (i) zero-shot activity recognition, (ii) unsupervised activity discovery, and (iii) unseen activity captioning. We show that the use of unpaired, multimodal data allows learning a shared embedding space that generalizes to unseen data. 

\begin{figure*}
    \begin{center}
      \includegraphics[width=0.95\textwidth]{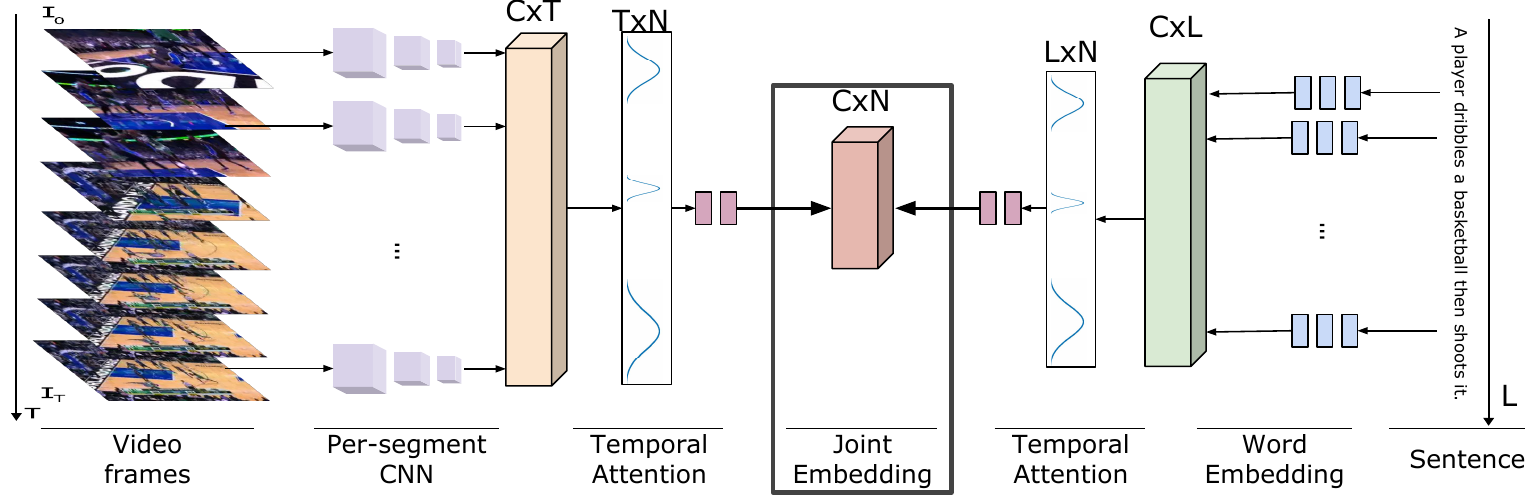}
    \end{center}
     \caption{Illustration of the encoder models used to learn a shared representation. Videos and sentences are mapped into a low-dimensional space by applying CNNs and temporal attention. Then several fully-connected layers map to the representation. The decoders follow this same architecture with the weights transposed.}
     \label{fig:model}  
\end{figure*}

\section{Related works}

\paragraph{Multimodal learning} Previous approaches to multimodal learning have used Restricted Boltzmann Machines \cite{srivastava2012multimodal} or log-bilinear models \cite{kiros2014multimodal} to learn distributions over sentences and images. Ngiam et al. \cite{ngiam2011multimodal} designed an autoencoder that learns joint audio-video representations, however relied on greedy, layer-by-layer training instead of training the model end-to-end. Similarly, Chandar et al. \cite{chandar2016correlational} proposed an auto-encoder able to learn correlations between different view of images. Frome et al. \cite{frome2013devise} describe a model that maps images and words to a shared embedding. However, these works either learn a joint embedding by concatenating the different features or require a triplet consisting of positive and negative pairs; they have not explored the use/effect of unpaired data.


\paragraph{Text and vision} Using both text and visual data has been studied for many tasks, such as image captioning~\cite{karpathy2014deep, johnson2016densecap, karpathy2015deep} or video captioning \cite{krishna2017dense, zhou2018end, xu2018joint}. Other works have explored the use of textual grounding for image/video retrieval~\cite{gupta2010using, rohrbach2016grounding, miech2017learning, anne2017localizing}. We note that using text for video retrieval/localization (e.g., \cite{anne2017localizing}) is similar in nature to the zero-shot or unseen recognition tasks. However, in those works, there is significant overlap between the text/video examples used in training and testing, while in our work we explicitly separate the classes used during training and evaluation; we focus on `unseen'.

There have been various models proposed to learn a fixed text embedding space with mappings from video features into this embedding space \cite{guadarrama2013youtube2text,otani2016learning, song2016unsupervised, wang2016learning, xian2016latent}. These works all learn a single directional mapping, without a shared representation space (which we find to be important). Further, most of them only learn with paired text/image samples and some require data in the form of positive/negative pairs. In this paper, we find learning a shared representation space and using unpaired, i.e., random additional data, to be important.

\paragraph{Learning with unpaired data} Recently, there have been many works taking advantage of variational autoencoders (VAEs) \cite{kingma2014auto} or generative adversarial networks (GANs) \cite{goodfellow2014generative} to learn mappings between unpaired samples. CycleGan~\cite{zhu2017unpaired} uses a cycle-consistency loss (i.e., the ability to go from a sample in one domain to a second domain then back to the source) to learn unpaired image translation (e.g., image to sketch). Other works learn many-to-many mappings between images \cite{almahairi2018augmented} or use two GANs to map between domains \cite{yi2017dualgan}. An autoencoder with shared weights for both domains has been used to learn a latent space for image-to-image translation \cite{liu2017unsupervised}. However, these works all focus on learning mappings between unpaired data of the same modalitiy (e.g. image to image), where the data is from the same underlying distribution. We focus on a more challenging problem: learning from different modalities with very different distributions, where we find directly using previous approaches do not perform well as they are.

\paragraph{Zero-shot activity recognition} There are works on zero-shot activity recognition. Common approaches include using attributes \cite{liu2011recognizing,palatucci2009zero,romera2015embarrassingly} or word embeddings \cite{xu2015semantic,xu2017transductive,norouzi2013zero,socher2013zero,kodirov2017semantic} or learning a similarity metric \cite{zhang2015zero,chopra2005learning}. Some works have explored using adversarial losses on the latent space \cite{chen2018zero}, used GANs to generate features for unseen classes \cite{xian2018feature}  or used auto-encoders \cite{wang2017zero}. Felix et al. \cite{felix2018multi} proposed a GAN-based approach to learn embeddings for zero-shot learning. Different from our approach, they applied the GAN only on the semantic, hand-crafted attributes of the classes to generate representations. We formulate a more general framework generating representations for all modalities, also taking advantage of more generic and challenging text and video.

Importantly, our work differs from these previous works in three key ways: (1) we show the benefit of using additional \textbf{unpaired} samples, (2) we experimentally compare the use of the representations for three tasks (i.e., zero-shot recognition, unseen recognition, and unseen video captioning), and (3) we learn a shared, multimodal representation with bi-directional mappings in an end-to-end fashion. We find that directly using the previous methods with unpaired data do not perform as well. 



\section{Method}
To enable learning of a shared representation, we use a deep autoencoder architecture. Our model consists of 4 neural networks:
\begin{align*}
\mbox{\textbf{Video Encoder} } & E_{V}: v \mapsto z_v
& 
\mbox{\textbf{Video Decoder} } & G_{V}: z \mapsto v
\\
\mbox{\textbf{Text Encoder} } & E_{T}: t \mapsto z_t
&
\mbox{\textbf{Text Decoder} } & G_{T}: z \mapsto t
\end{align*}
where $v$ is a sequence of video data and $t$ is a sentence (sequence of words). $z$ is the representation in the shared space that we are learning. The encoders learn a compressed representation of the video or text while the decoders are trained to reconstruct the input:
\begin{equation}
\label{eq:recons}
    \mathcal{L}_{recons}(v,t) = ||G_V(E_V(v)) - v||_2 + ||G_T(E_T(t)) - t||_2
\end{equation}
As both text and video data are sequences, they often have different lengths. A shared representation requires that the features from both modalities have the same dimensions.
Given a text representation of length $L$ and a video representation of length $T$, we need to obtain a fixed-size representation. 
To learn a fixed-dimensional representation, there are many choices for the encoder/decoder architecture, such as temporal pooling \cite{ng2015beyond}, attention \cite{piergiovanni2017learning} or RNNs \cite{krishna2017dense}. 
We chose temporal attention filters \cite{piergiovanni2017learning} as they learn a mapping from any length input to a $N$-dimensional vector and have been shown to outperform temporal pooling and RNNs on activity recognition tasks.  

The attention filters consists of $N$ Gaussians, each learning 2 parameters: a center $\hat{g}$ and width $\sigma$, which are constrained to be positive. The filters are determined by:
\begin{equation} \label{eq:mrf_match}
\begin{split}
g_n &= 0.5\cdot T \cdot (\hat{g}_n + 1)\\
F[n,t] &= \frac{1}{Z} \exp(-\frac{(t-g_n)^2}{2\sigma_n^2})\\
&n\in\{0,1,\ldots,N-1\},~ t\in\{0,1,\ldots,T-1\} \\
\end{split}
\end{equation}
The weights are applied by matrix multiplication with the video or text sequence (e.g., the outputs of $E_V$ or $E_T$): $v' = Fv$. This (i.e., $v'$) is then used as the representations for the joint space. Additionally, we can learn a transposed version of these filters to reconstruct the input: $v=F^Tv'$. To reconstruct the input, the decoders learn their own parameters with the tensors transposed, resulting in the matching output size. Fig.~\ref{fig:model} shows our encoder architecture.


\subsection{Learning a joint embedding space}
\begin{figure*}
    \centering
      \includegraphics[width=0.95\textwidth]{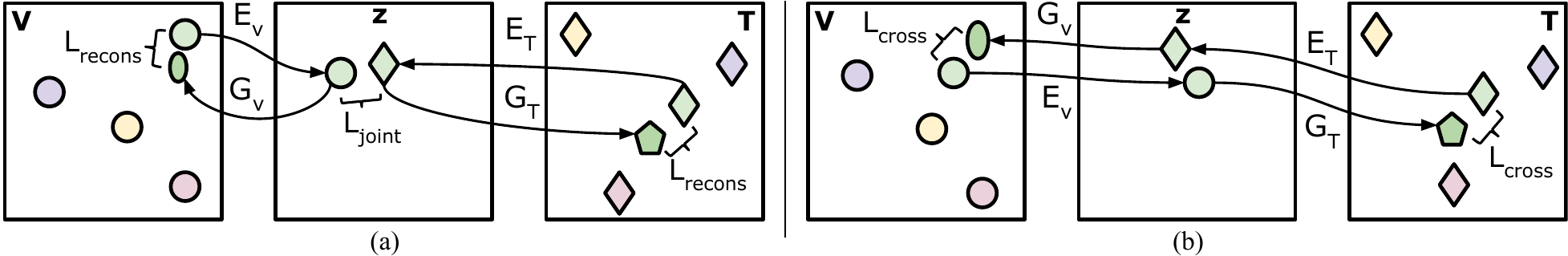}
      \caption{Visualization of several constrains on the shared embedding space. Circles are video data, ovals are reconstructed video. Diamonds are text data, and pentagons are reconstructed text. \textbf{(a)} The reconstruction (Eq. \ref{eq:recons}) and joint (Eq. \ref{eq:joint}) losses. \textbf{(b)} Mapping from text to video using the cross-domain (Eq. \ref{eq:cross}) loss.}
      \label{fig:paired-data}
\end{figure*}

To learn a joint representation space, we minimize the $L_2$ distance between the embeddings of a pair of text and video (shown in Fig. \ref{fig:paired-data}(a)):
\begin{equation}
\label{eq:joint}
    \mathcal{L}_{joint}(v,t) = ||E_V(v) - E_T(t)||_2
\end{equation}
This forces the joint embeddings to be similar and when combined with the reconstruction loss, ensures that the representations can still reconstruct the input.

We can further constrain the networks and learned representation by forcing a cross-domain mapping from text to video and from video to text (shown in Fig. \ref{fig:paired-data}(b)):
\begin{equation}
\label{eq:cross}
    \mathcal{L}_{cross}(v,t) = ||G_T(E_V(v)) - t||_2 + ||G_V(E_T(t)) - v||_2
\end{equation}

Additionally, we can use a `cycle' loss to map from video to text and back to video. Note that while the previous losses all require paired examples, this loss does not.
\begin{equation}
\label{eq:cycle}
\begin{split}
    \mathcal{L}_{cycle}(v,t) &= ||G_T(E_V(G_V(E_T(t)))) - t||_2\\
    &+ ||G_V(E_T(G_T(E_V(v)))) - v||_2
\end{split}
\end{equation}

To train the model to learn a joint embedding space, we minimize
\begin{equation}
\label{eq:paired}
\begin{split}
\mathcal{L}(v,t) &= \mathcal{L}_{recons}(v,t) + \alpha_1\mathcal{L}_{joint}(v,t)  \\
&+ \alpha_2\mathcal{L}_{cross}(v,t) + \alpha_3\mathcal{L}_{cycle}(v,t)
\end{split}
\end{equation}
where $\alpha_i$ are hyper-parameters weighting the various loss components.

\subsection{Semi-supervised learning with unpaired data}
\begin{figure}
\begin{minipage}{0.48\textwidth}
    \centering
      \includegraphics[width=\linewidth]{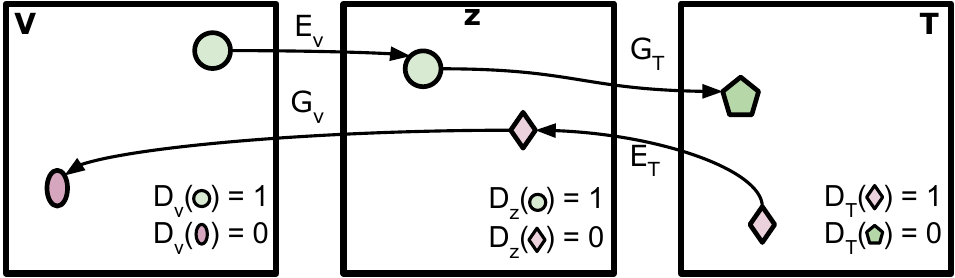}
      \caption{Visualization of the adversarial formulation to learn with unpaired data. We create 3 discriminators, (1) $D_z$ learns to discriminate examples of text/video in the latent space. (2) $D_V$ learns to discriminate video generated from text compared to video. (3) $D_T$ learns to discriminate generated text compared to text.}
      \label{fig:unpaired-data}
\end{minipage}
\hfill
\begin{minipage}{0.48\textwidth}
\begin{algorithm}[H]  
  \caption{Semi-supervised alignment with adversarial learning
    \label{alg:train}}  
  \begin{algorithmic}  
    \Function{Train}{}
    \For{number of initialization iterations}
       \State Sample ($V$, $T$) from paired training data
        \State Update encoders/decoders (Eq. \ref{eq:paired}) 
        \State Update discriminators (Eq. \ref{eq:discrim})
    \EndFor
    \For{number of training iterations}
        \State Sample $P=$($V_p$, $T_p$) from paired and
        \State $U=$($V_u$, $T_u$) from unpaired training data
        \State Update encoders/decoders with $P$ (Eq. \ref{eq:paired})
        \State Update encoders/decoders with $U$ (Eq. \ref{eq:gen})
        \State Update discriminators based on all (Eq. \ref{eq:discrim})
    \EndFor
    \EndFunction
  \end{algorithmic}  
\end{algorithm}
\end{minipage}
\end{figure}

To learn using unpaired data (i.e., unrelated text and video), we use an adversarial formulation. We treat the encoders and decoders as generator networks. We then learn an additional 3 discriminator networks which constrain the generators and embedding space and force the encoders and decoders to be consistent:
\begin{enumerate}[(1)]
    \item $D_z$ which learns to discriminate between latent text representations and latent video representations. Conceptually, this constrains the learned embeddings to appear to be from the same distribution.
    \item  $D_V$ which learns to discriminate between true video data and generated video data $G_V(E_T(t))$.
    \item $D_T$ which learns to discriminate between true text data and generated text data, $G_T(E_V(v))$. 
\end{enumerate}

Given these discriminators, we minimize the following losses:
\begin{equation}
\label{eq:discrim}
\begin{split}
    \loss_{D_z}(v,t) &= - \log (D_z(E_T(t)))  - \log (1 -  D_z(E_V(v)))\\
    \loss_{D_V}(v,t) &= -\log (D_V(v)  - \log (1 -  D_V(G_V(E_T(t))))\\
    \loss_{D_T}(v,t) &= -\log (D_T(t))  - \log (1 -  D_T(G_T(E_V(v))))\\
\end{split}
\end{equation}

Using the discriminators, we can train the generators (encoders and decoders) to minimize the following loss based on unpaired data:
\begin{equation}
\label{eq:gen}
\begin{split}
    \loss_{G_z}(v,t) &= \log (D_z(E_T(t))))  + \log (1 -  D_z(E_V(v)))\\
    \loss_{G_V}(v,t) &= \log (1 -  D_V(G_V(E_T(t))))\\
    \loss_{G_T}(v,t) &= \log (1 -  D_T(G_T(E_V(v))))\\
\end{split}
\end{equation}
Note that in this formulation, $v$ and $t$ are not paired. 

These networks are trained in an adversarial setting. For example, for the text-to-video generator (i.e., $v' = G_V(E_T(t))$ and video discriminator, $D_V$, we optimize the following minimax equation:
\begin{equation}
\begin{split}
    \min_{E_T,G_V} & \max_{D_V} = \E_{v\sim p_{\text{data}}(v)} [\log D_V(v)] \\
    &+ \E_{t\sim p_{\text{data}}(t)} [\log (1-D_V(G_V(E_T(t))))]
\end{split}
\end{equation}

This equation is similarly applied for video-to-text. For learning the embedding space with the video and text encoders, $E_V,E_T$ and the discriminator $D_z$, we optimize the following minimax equation:
\begin{equation}
\begin{split}
    \min_{E_T,E_V} & \max_{D_z} = \E_{v\sim p_{\text{data}}(v)} [\log D_z(E_V(v))] \\
    &+ \E_{t\sim p_{\text{data}}(t)} [\log (1-D_z(E_T(t)))]
\end{split}
\end{equation}

As training GANs can be unstable, we developed a method to allow for more stable training of the joint embedding space, shown in Algorithm~\ref{alg:train}. We initialize both the generator and discriminator networks by training only on paired data. After several iterations of this, we train with both unpaired and paired data. We found the initial training of the generators and discriminators was important for stability, without it the loss often diverges and the learned embedding did not generalize to unseen activities.

\begin{table*}
  \caption{Comparison of accuracy of various methods on ActivityNet for 5, 10, 20 or 50 unseen classes. These results are averaged over 10 trials where each trial has a different set of unseen activities.}
  \label{res:actnet}
  \centering
  \renewcommand{\arraystretch}{0.9}
  \begin{tabular}{lcccc}
    \toprule
         &  5 Unseen     & 10 Unseen & 20 Unseen & 50 Unseen \\
    \midrule
    Paired Data \\
    \midrule
    Fixed Text Representation       & 41.9 & 38.4 & 29.4 & 15.6\\
    Triplet Loss     & 56.8     & 44.9    & 38.8    & 23.3  \\
    joint            & 54.3 & 41.7 & 36.1 & 21.2 \\
    recons + cross                  & 21.1 & 12.6 & 7.6  & 2.9 \\
    joint + recons  & 70.1 & 54.4 & 42.6 & 27.5\\
    joint + recons + cycle  & 70.4 & 54.3 & 42.1 & 26.8\\
    joint + recons + cross  & 72.6 & 55.4 & 43.2 & 27.8\\
    joint + recons + cross + cycle                    & 76.4 &  56.9  & 45.5 & 28.8\\
    triplet + recons + cross + cycle                    & 76.7 &  57.2  & 46.3 & 29.1\\
    
    \midrule
    \multicolumn{5}{l}{With Adversarial Losses (triplet + recons + cross + cycle + Adv.)}\\
    \midrule
    + $D_z$                    & 78.5 &  57.4  & 45.9 & 29.3\\
    + $D_v + D_t$                    & 77.4 &  57.2  & 45.7 & 28.9\\
    + $D_z + D_v + D_t$                    & 79.8 &  58.4  & 46.5 & 29.8\\

    \midrule
    Paired + Unpaired Data \\
    \midrule
    recons + cycle  & 22.8 & 13.6 & 8.4 &  4.2 \\
    triplet + recons + cycle & 72.6 & 58.4 & 44.7 & 29.3 \\
    triplet + recons + cross + cycle & 73.4 & 59.1 & 45.3 & 29.2 \\
    Without Algorithm \ref{alg:train} & 23.4 & 11.7 & 6.5 & 3.1\\
    All terms       & 82.5 & 60.4 & 46.2 & 30.1\\
    \bottomrule
  \end{tabular}
\end{table*}

\section{Experiments}
We compare our various approaches on different tasks (i) zero-shot activity recognition, (ii) unsupervised activity discovery and (iii) unseen activity captioning. These tasks test various combinations of our encoders and decoders and how well the shared representation generalizes to unseen data. We experimentally confirm the benefits of our methods using multiple public datasets: AcitivtyNet \cite{heilbron2015activitynet,krishna2017dense}, HMDB \cite{kuehne2011hmdb}, UCF101 \cite{soomro2012ucf101}, and MLB-YouTube \cite{mlbyoutube2018}. Implementation details can be found in the Appendix.

\paragraph{Baselines} For baselines, we compare to a fixed-text embedding space, were only a mapping from video data into the text embedding space is learned (e.g., \cite{otani2016learning}). We also compare to learning a shared embedding space with the `recons' (Eq. \ref{eq:recons}) and `cross' (Eq. \ref{eq:cross}) terms (e.g., \cite{ngiam2011multimodal}). We additionally compare to methods like CycleGan~\cite{zhu2017unpaired}, using various components without Algorithm \ref{alg:train}.

\subsection{Zero-shot activity recognition}

Zero-shot activity recognition is the problem of classifying a video that belongs to a class not seen during training. Given training videos of seen classes together with paired text descriptions, our approach learns a shared embedding that maps videos/texts from multiple seen classes. The objective is to classify videos of unseen classes solely based on the learned embedding space and the text samples.

To enable recognition of unseen activities, we use a sentence of the new, unseen class and obtain its representation in the shared space. We can then obtain representations of videos in the same space, using nearest neighbors matching to classify each clip. Such approach takes advantage of the learned textual relationships (e.g., \cite{mikolov2013efficient}) and the shared, multimodal representation space.

We use the ActivityNet captions \cite{krishna2017dense} dataset to learn the shared representations, as this dataset has both sentence descriptions for each video as well as activity classes. We randomly choose a set of $K$ activity classes and withhold all videos/sentences belonging to those classes during training. For evaluating on the unseen activities, we take a subset of sentences for the unseen classes and map the sentences into the joint embedding space, $z_t = E_T(t)$. We then map the videos into the space, $z_v = E_V(v)$ and use nearest neighbors to match each video ($z_v$) to text ($z_t$), using the class of the nearest sentence as the classification for the video. We rely on the similarities between the representations (e.g., word embeddings) to enable the models ability to generalize to these unseen classes.

In Table~\ref{res:actnet}, we compare the effect of the various loss components. For each method, we run 10 trials each with a different set of unseen activity classes and average the results. We find that previous methods of learning a fixed language embedding (e.g., \cite{socher2013zero,xu2015semantic,xu2017transductive}) are significantly outperformed by learning a joint representation. Previous methods learning embedding spaces without the `joint' term (e.g. \cite{ngiam2011multimodal}), we found yield nearly random performance on these tasks, suggesting that forcing the representations to match in the embedding space is important. Further, adding the reconstruction, cross-domain, and cycle losses all improve performance. We also compare to a standard triplet loss (e.g., \cite{guadarrama2013youtube2text}) which requires positive/negative samples. We find that the triplet loss outperforms the `joint' loss, but is surpassed by adding the `cycle' and `cross' terms, which use less data. We also compared using the triplet loss when combined with the other terms, finding a slight improvement over the joint term. Note using both the joint and triplet would be redundant, since the triplet loss contains the joint loss terms.

We also compare the various components of the adversarial loss. We compare to having just the adversarial loss on the representation ($D_z$), like \cite{chen2018zero}, and compare just the adversary on the generated videos/sentences. We find the use of all terms is important for performance. 

While previous works such as \cite{ngiam2011multimodal} can support learning with unpaired data, we find that the adversarial loss provides better results than just the `cycle' and `recons' terms, and further improves over training with just paired data. Further, we find that CycleGan-style approaches, without Algorithm \ref{alg:train}, fail on this task.

In Table~\ref{res:previous}, we compare our approach to previous zero-shot learning methods on HMDB and UCF101. The paired training data for these models is drawn from ActivityNet with any classes belonging to HMDB or UCF101 withheld. The unpaired text data is sampled from Charades and the video data comes from either HMDB (when testing on UCF101) or UCF101 (when testing on HMDB). As HMDB and UCF101 have no text descriptions, we created a sentence description for each activity class (included in Appendix B). We find that the shared representation outperforms the previous approaches on these datasets and unpaired adversarial learning further improves performance.

\begin{table}
  \caption{Results on HMDB51 and UCF101 (accuracy) compared to previous state-of-the-art results. We find that learning a shared representation is beneficial and that augmented with unpaired data provides the best results.}
  \label{res:previous}
  \centering
  \renewcommand{\arraystretch}{0.9}
  \small
  \begin{tabular}{lccccc}
    \toprule
         & Feat & HMDB51  & UCF101 \\
    \midrule
    SJE \cite{akata2015evaluation} & IDT & $12.0\pm 2.6$ & $9.3 \pm 1.7$\\
    ConSe \cite{norouzi2013zero} & IDT  & $15.0 \pm 2.7$ & $11.6\pm 2.1$\\
    ZSECOC~\cite{qin2017zero} & IDT   & $22.6\pm 1.2$  & $15.1\pm 1.7$ \\
    SE~\cite{xu2015semantic} & IDT      & $21.2\pm 3.0$  & $18.6\pm 2.2$ \\
    MRR~\cite{xu2017transductive} & IDT & $24.1\pm 3.8$  & $22.1\pm 2.5$\\
    SAE \cite{kodirov2017semantic} & I3D  & $25.6\pm 3.2$  & $25.4\pm 2.2$\\
    Ours (paired) & IDT      & $26.3 \pm 3.2$ & $25.4 \pm 3.4$  \\
    Ours (paired + unpaired)     & IDT          & $29.7 \pm 2.2$ & $26.4 \pm 2.1$ \\
    Ours (paired) & I3D      & $28.3 \pm 2.7$ & $27.8 \pm 2.2$  \\
    Ours (paired + unpaired)     & I3D          & $34.7 \pm 2.4$ & $33.4 \pm 1.8$ \\
    \bottomrule
  \end{tabular}
\end{table}

\begin{table}
  \caption{Comparison of various source of unpaired data on ActivityNet with 10 unseen classes, values reported for both unseen classes and all (seen+unseen) classes. Results are accuracy, higher is better.}
  \label{res:unpaired-sources}
  \centering
  \renewcommand{\arraystretch}{0.9}
  \begin{tabular}{lcc}
    \toprule
         &  Unseen & All \\
    \midrule
    Paired Data                 & 58.3 & 69.6   \\
    + Random Wikipedia Sentences  & 55.8 & 66.4   \\
    + Random Dictionary Defs.     & 56.3 & 68.2 \\
    + Verb Dictionary Defs. & 59.2 & 70.7 \\
    + Random YouTube Videos         & 58.7 & 70.1  \\
    + Verbs + Random Videos  & 60.3 & 71.2 \\
    \bottomrule
  \end{tabular}
\end{table}

\begin{table}
  \centering
    \caption{Comparison of unsupervised activity classification on MLB-YouTube.}
  \renewcommand{\arraystretch}{0.9}
  \begin{tabular}{lcc}
    \toprule
         &  Accuracy & mAP \\
    \midrule
    Baseline I3D features & 23.4 & 32.6 \\
    Fixed Text Representation       & 27.9 & 34.7 \\
    joint            & 34.5 & 41.6 \\
    joint + recons  & 37.9 & 43.7 \\
    joint + recons + cycle  & 44.2 & 48.6 \\
    joint + recons + cross  & 43.7  & 49.3 \\
    triplet + recons + cross  & 43.9  & 49.5 \\
    All (paired)                    & 48.4 & 51.2  \\
    All (+ unrelated unpaired)      & 39.7 & 43.9  \\
    All (+ related unpaired)      & 49.1  & 54.3 \\
    \bottomrule
  \end{tabular}
  \label{res:mlb}
\end{table}

\begin{table*}
  \caption{Unseen activity recognition results (accuracy) on ActivityNet, HMDB51 and UCF101, evaluated by using both unseen and seen classes for the testing.}
  \label{res:full-sets}
  \centering
  \setlength{\tabcolsep}{6pt}
  \renewcommand{\arraystretch}{0.9}
  \begin{tabular}{lcccc}
    \toprule
         & ActNet (10 unseen) & ActNet (50 unseen)  & HMDB51  & UCF101 \\
    \midrule
    Fixed Text Representation   & 55.7 & 46.8 & 24.5 & 26.8 \\
    Triplet Loss                & 57.7 & 48.5 & 27.6  & 29.8  \\
    joint                       & 62.1 & 50.2 & 29.8 & 30.6 \\
    joint + recons              & 64.4 & 52.6 & 30.4 & 31.3 \\
    joint + recons + cross + cycle  & 69.6 & 58.5  & 35.6 & 36.5 \\
    triplet + recons + cross + cycle  & 69.8 & 58.6  & 35.7 & 36.8 \\
    \midrule
    \multicolumn{5}{l}{Paired + Unpaired Data} \\
    \midrule
    All terms       & 71.7 & 65.9 & 38.9 & 42.2 \\
    \bottomrule
  \end{tabular}
\end{table*}

\subsection{Use of Unpaired Data}
We explore different strategies for obtaining unpaired data. Keeping a fixed set of paired text and videos, we explore adding various sources of unpaired data: (i) 10k random Wikipedia sentences, (ii) 10k random dictionary definitions, and (iii) 10k verb dictionary definitions. We also compare adding 10k random videos from YouTube as additional video data. Ours results using 10 unseen classes are in Table \ref{res:unpaired-sources}. We find that augmenting with similar unpaired data improves performance, while irrelevant data harms performance. We find that dictionary verb definitions improve performance the most, as they capture important semantic information regarding the activities we are learning. The use of additional video data is further beneficial.

\subsection{Unsupervised activity discovery}

\begin{figure*}
    \centering
      \includegraphics[width=\linewidth]{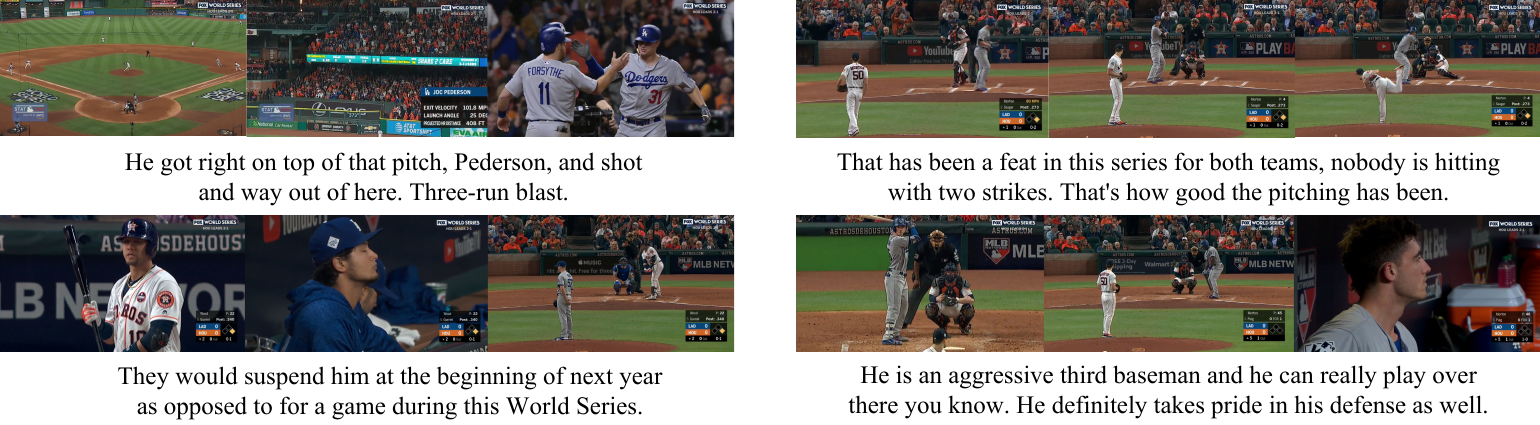}
      \caption{Example video sequences from the MLB-YouTube dataset with the commentary caption. \textbf{Top}: Sentences that describe the occurring activities. \textbf{Bottom}: Sentences that do not describe the current activities.}
      \label{fig:baseball-ex-caption}
\end{figure*}

To further evaluate the shared representation, we conducted experiments on unsupervised activity discovery. For this task, we expanded the MLB-YouTube dataset \cite{mlbyoutube2018} by densely annotating the videos with a transcription of the announcers' commentary, resulting in approximately 50 hours of aligned text and video. Examples of this data are shown in Fig. \ref{fig:baseball-ex-caption}. The MLB-YouTube dataset is designed for fine-grained activity recognition, where the difference between activities is quite small. Additionally, these captions only roughly describe what is happening in the video, and often contain unrelated stories or commentary on a previous event, making this a challenging task. 
The dataset will be made publicly available.  
To train the shared representation, we split each baseball video into 30 second intervals and use the corresponding text as paired data, resulting in 6,089 paired training samples.

\begin{figure*}
    \centering
      \includegraphics[width=\textwidth]{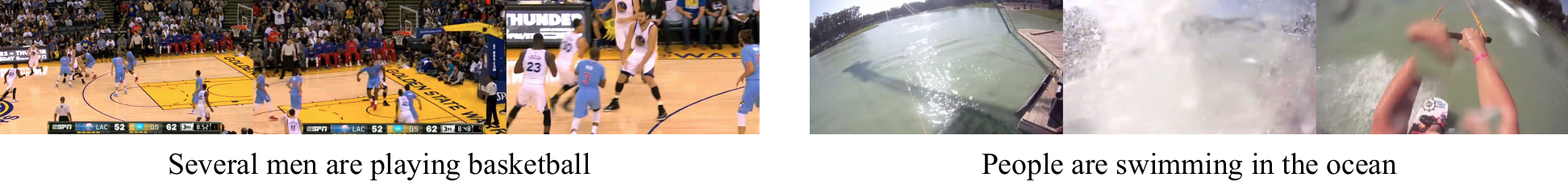}
      \caption{Example captions for unseen activities. \textbf{Left}: Using a shared representation allows the model to correctly caption this video as basketball, despite never seeing an example of basketball during training. \textbf{Right}: An example of a caption for the unseen water-ski activity. Here the model fails to correctly caption the activity.}
      \label{fig:unseen-caps}
\end{figure*}

We evaluate the shared representation using the segmented videos from MLB-YouTube. For each video, we compute the embedded features and apply $k$-means clustering ($k=8$, the number of classes). Each segmented video is assigned to a cluster and votes for the cluster label based on its ground truth label. We use that cluster assignment for classification on the MLB-YouTube test set. We report our findings in Table~\ref{res:mlb}. As a baseline, we cluster I3D features pre-trained on Kinetics. We find that our methods improve the representation. However, we note that when using unpaired data from Charades, the performance drops. This is likely due to Charades data being very different from MLB-YouTube data. We collected additional captions and baseball videos to augment the MLB-YouTube dataset, and confirmed that unpaired data helps when it is from a similar distribution.


In Table~\ref{res:unsupervised} we compare various methods for unsupervised activity discovery on HMDB and UCF101. Here, we learn a shared representation using the ActivityNet videos and captions. We withhold any videos belonging to a class in HMDB or UCF101. Unlike MLB-YouTube, on these datasets, we find that using the unpaired training with Charades further improves performance. This confirms that when the additional data is similar to the target dataset, using the adversarial learning setting further improves the representations.

\begin{table}
  \centering
    \caption{Comparison of unsupervised activity classification on HMDB and UCF101.}
  \begin{tabular}{lcc}
    \toprule
         &  HMDB & UCF101 \\
    \midrule
    I3D features           & 26.6 & 42.5 \\
    Joint            & 32.4 & 57.7 \\
    Joint + recons                   & 33.5  & 59.0 \\
    All (paired)           &  34.6 & 59.5 \\
    All (+ unpaired)             & 34.9 & 59.9 \\
    \bottomrule
  \end{tabular}
  \label{res:unsupervised}
\end{table}

\subsection{Unseen video captioning}

As our model learns a bi-directional mappings, we can apply our model to generate video captions. Existing video captioning models are unable to create realistic captions for unseen activities, as without training data they do not know the words to describe the video. Given a video, $v$, we can generate a caption by mapping the video to text $t=G_T(E_V(v))$. For each word, we then use nearest neighbors matching with the GloVe embeddings to obtain the words to form a sentence. We find that using our method with paired and unpaired data improves performance using METEOR (3.6 to 6.9) \cite{banerjee2005meteor} and CIDEr \cite{vedantam2015cider} (8.9 to 13.9) scores. For these metrics, higher values are better and are measured with the unseen classes from the ActivityNet dataset.
In Table~\ref{res:caption}, we report the commonly used METEOR \cite{banerjee2005meteor} and CIDEr \cite{vedantam2015cider} scores of our various models, measured with the unseen classes from the ActivityNet dataset. 
We find that learning a shared representation (4.1) is beneficial and using unpaired samples further improves the task (5.3 paired only vs 6.9 paired and unpaired). In Fig. \ref{fig:unseen-caps}, we show example captioned videos. Note that this task is extremely challenging, as it requires the model to generate captions using activity words (e.g., basketball) not seen during training.

\begin{table}
  \caption{Comparison of several models for unseen activity captioning using the ActivityNet dataset, using METEOR and CIDEr scores. This evaluation was done on 10 unseen classes held out during training. Higher values are better.}
  \label{res:caption}
  \centering
  \begin{tabular}{lcc}
    \toprule
         &  METEOR & CIDEr \\
    \midrule
    Fixed Text Representation       & 3.64  & 8.95   \\
    Joint                           & 4.21  & 9.23   \\
    All (paired)                    & 5.31  & 11.21   \\
    All (paired + unpaired)         & 6.89  & 13.95   \\
    \bottomrule
  \end{tabular}
\end{table}

\section{Conclusion}
We proposed an approach to learn a joint language/text representation using various constraints. We further extended the model to be able to learn with unpaired video and text data using an adversarial formulation. We experimentally confirmed that learning with unpaired data is beneficial to three difficult tasks (i) zero-shot activity classification, (ii) unsupervised activity discovery, and (iii) unseen activity captioning. We find that the use of related unpaired data is beneficial. We presented several strategies for obtaining unpaired data and confirmed the benefit of adding additional, relevant unpaired data.

\paragraph{Acknowledgement} This work was supported in part by the National Science Foundation (IIS-1812943 and CNS-1814985).

{\small
\bibliography{bib}

\begin{thebibliography}{10}\itemsep=-1pt

\bibitem{akata2015evaluation}
Z.~Akata, S.~Reed, D.~Walter, H.~Lee, and B.~Schiele.
\newblock Evaluation of output embeddings for fine-grained image
  classification.
\newblock In {\em Proceedings of the IEEE Conference on Computer Vision and
  Pattern Recognition (CVPR)}, 2015.

\bibitem{almahairi2018augmented}
A.~Almahairi, S.~Rajeswar, A.~Sordoni, P.~Bachman, and A.~Courville.
\newblock Augmented cyclegan: Learning many-to-many mappings from unpaired
  data.
\newblock {\em arXiv preprint arXiv:1802.10151}, 2018.

\bibitem{banerjee2005meteor}
S.~Banerjee and A.~Lavie.
\newblock Meteor: An automatic metric for mt evaluation with improved
  correlation with human judgments.
\newblock In {\em Proceedings of the acl workshop on intrinsic and extrinsic
  evaluation measures for machine translation and/or summarization}, 2005.

\bibitem{carreira2017quo}
J.~Carreira and A.~Zisserman.
\newblock Quo vadis, action recognition? a new model and the kinetics dataset.
\newblock In {\em Proceedings of the IEEE Conference on Computer Vision and
  Pattern Recognition (CVPR)}, 2017.

\bibitem{chandar2016correlational}
S.~Chandar, M.~M. Khapra, H.~Larochelle, and B.~Ravindran.
\newblock Correlational neural networks.
\newblock {\em Neural computation}, 28(2):257--285, 2016.

\bibitem{chen2018zero}
L.~Chen, H.~Zhang, J.~Xiao, W.~Liu, and S.-F. Chang.
\newblock Zero-shot visual recognition using semantics-preserving adversarial
  embedding network.
\newblock In {\em Proceedings of the IEEE Conference on Computer Vision and
  Pattern Recognition (CVPR)}, 2018.

\bibitem{chopra2005learning}
S.~Chopra, R.~Hadsell, and Y.~LeCun.
\newblock Learning a similarity metric discriminatively, with application to
  face verification.
\newblock In {\em Proceedings of the IEEE Conference on Computer Vision and
  Pattern Recognition (CVPR)}. IEEE, 2005.

\bibitem{felix2018multi}
R.~Felix, V.~B. Kumar, I.~Reid, and G.~Carneiro.
\newblock Multi-modal cycle-consistent generalized zero-shot learning.
\newblock In {\em Proceedings of European Conference on Computer Vision
  (ECCV)}, pages 21--37, 2018.

\bibitem{frome2013devise}
A.~Frome, G.~S. Corrado, J.~Shlens, S.~Bengio, J.~Dean, T.~Mikolov, et~al.
\newblock Devise: A deep visual-semantic embedding model.
\newblock In {\em Advances in Neural Information Processing Systems (NIPS)},
  2013.

\bibitem{goodfellow2014generative}
I.~Goodfellow, J.~Pouget-Abadie, M.~Mirza, B.~Xu, D.~Warde-Farley, S.~Ozair,
  A.~Courville, and Y.~Bengio.
\newblock Generative adversarial nets.
\newblock In {\em Advances in Neural Information Processing Systems (NIPS)},
  2014.

\bibitem{guadarrama2013youtube2text}
S.~Guadarrama, N.~Krishnamoorthy, G.~Malkarnenkar, S.~Venugopalan, R.~Mooney,
  T.~Darrell, and K.~Saenko.
\newblock Youtube2text: Recognizing and describing arbitrary activities using
  semantic hierarchies and zero-shot recognition.
\newblock In {\em Proceedings of the IEEE International Conference on Computer
  Vision (ICCV)}. IEEE, 2013.

\bibitem{gupta2010using}
S.~Gupta and R.~J. Mooney.
\newblock Using closed captions as supervision for video activity recognition.
\newblock In {\em Proceedings of the American Association for Artificial
  Intelligence (AAAI)}, 2010.

\bibitem{heilbron2015activitynet}
F.~C. Heilbron, V.~Escorcia, B.~Ghanem, and J.~C. Niebles.
\newblock Activitynet: A large-scale video benchmark for human activity
  understanding.
\newblock In {\em Proceedings of the IEEE Conference on Computer Vision and
  Pattern Recognition (CVPR)}. IEEE, 2015.

\bibitem{anne2017localizing}
L.~Hendricks, O.~Wang, E.~Shechtman, J.~Sivic, T.~Darrell, and B.~Russell.
\newblock Localizing moments in video with natural language.
\newblock In {\em Proceedings of the IEEE Conference on Computer Vision and
  Pattern Recognition (CVPR)}, 2017.

\bibitem{johnson2016densecap}
J.~Johnson, A.~Karpathy, and L.~Fei-Fei.
\newblock Densecap: Fully convolutional localization networks for dense
  captioning.
\newblock In {\em Proceedings of the IEEE Conference on Computer Vision and
  Pattern Recognition (CVPR)}, 2016.

\bibitem{karpathy2015deep}
A.~Karpathy and L.~Fei-Fei.
\newblock Deep visual-semantic alignments for generating image descriptions.
\newblock In {\em Proceedings of the IEEE Conference on Computer Vision and
  Pattern Recognition (CVPR)}, 2015.

\bibitem{karpathy2014deep}
A.~Karpathy, A.~Joulin, and L.~F. Fei-Fei.
\newblock Deep fragment embeddings for bidirectional image sentence mapping.
\newblock In {\em Advances in Neural Information Processing Systems (NIPS)},
  2014.

\bibitem{kingma2014auto}
D.~P. Kingma and M.~Welling.
\newblock Auto-encoding variational bayes.
\newblock {\em International Conference on Learning Representations (ICLR)},
  2014.

\bibitem{kiros2014multimodal}
R.~Kiros, R.~Salakhutdinov, and R.~Zemel.
\newblock Multimodal neural language models.
\newblock In {\em International Conference on Machine Learning (ICML)}, 2014.

\bibitem{kodirov2017semantic}
E.~Kodirov, T.~Xiang, and S.~Gong.
\newblock Semantic autoencoder for zero-shot learning.
\newblock {\em arXiv preprint arXiv:1704.08345}, 2017.

\bibitem{krishna2017dense}
R.~Krishna, K.~Hata, F.~Ren, L.~Fei-Fei, and J.~C. Niebles.
\newblock Dense-captioning events in videos.
\newblock In {\em Proceedings of the IEEE Conference on Computer Vision and
  Pattern Recognition (CVPR)}, 2017.

\bibitem{kuehne2011hmdb}
H.~Kuehne, H.~Jhuang, E.~Garrote, T.~Poggio, and T.~Serre.
\newblock Hmdb: a large video database for human motion recognition.
\newblock In {\em Proceedings of the IEEE International Conference on Computer
  Vision (ICCV)}, 2011.

\bibitem{liu2011recognizing}
J.~Liu, B.~Kuipers, and S.~Savarese.
\newblock Recognizing human actions by attributes.
\newblock In {\em Proceedings of the IEEE Conference on Computer Vision and
  Pattern Recognition (CVPR)}. IEEE, 2011.

\bibitem{liu2017unsupervised}
M.-Y. Liu, T.~Breuel, and J.~Kautz.
\newblock Unsupervised image-to-image translation networks.
\newblock In {\em Advances in Neural Information Processing Systems (NIPS)},
  2017.

\bibitem{miech2017learning}
A.~Miech, J.-B. Alayrac, P.~Bojanowski, I.~Laptev, and J.~Sivic.
\newblock Learning from video and text via large-scale discriminative
  clustering.
\newblock In {\em Proceedings of the IEEE International Conference on Computer
  Vision (ICCV)}, 2017.

\bibitem{mikolov2013efficient}
T.~Mikolov, K.~Chen, G.~Corrado, and J.~Dean.
\newblock Efficient estimation of word representations in vector space.
\newblock {\em arXiv preprint arXiv:1301.3781}, 2013.

\bibitem{ng2015beyond}
J.~Y.-H. Ng, M.~Hausknecht, S.~Vijayanarasimhan, O.~Vinyals, R.~Monga, and
  G.~Toderici.
\newblock Beyond short snippets: Deep networks for video classification.
\newblock In {\em Proceedings of the IEEE Conference on Computer Vision and
  Pattern Recognition (CVPR)}, 2015.

\bibitem{ngiam2011multimodal}
J.~Ngiam, A.~Khosla, M.~Kim, J.~Nam, H.~Lee, and A.~Y. Ng.
\newblock Multimodal deep learning.
\newblock In {\em International Conference on Machine Learning (ICML)}, 2011.

\bibitem{norouzi2013zero}
M.~Norouzi, T.~Mikolov, S.~Bengio, Y.~Singer, J.~Shlens, A.~Frome, G.~S.
  Corrado, and J.~Dean.
\newblock Zero-shot learning by convex combination of semantic embeddings.
\newblock {\em arXiv preprint arXiv:1312.5650}, 2013.

\bibitem{otani2016learning}
M.~Otani, Y.~Nakashima, E.~Rahtu, J.~Heikkil{\"a}, and N.~Yokoya.
\newblock Learning joint representations of videos and sentences with web image
  search.
\newblock In {\em Proceedings of European Conference on Computer Vision
  (ECCV)}, 2016.

\bibitem{palatucci2009zero}
M.~Palatucci, D.~Pomerleau, G.~E. Hinton, and T.~M. Mitchell.
\newblock Zero-shot learning with semantic output codes.
\newblock In {\em Advances in Neural Information Processing Systems (NIPS)},
  2009.

\bibitem{pennington2014glove}
J.~Pennington, R.~Socher, and C.~D. Manning.
\newblock Glove: Global vectors for word representation.
\newblock In {\em Empirical Methods in Natural Language Processing (EMNLP)},
  2014.

\bibitem{piergiovanni2017learning}
A.~Piergiovanni, C.~Fan, and M.~S. Ryoo.
\newblock Learning latent sub-events in activity videos using temporal
  attention filters.
\newblock In {\em Proceedings of the American Association for Artificial
  Intelligence (AAAI)}, 2017.

\bibitem{mlbyoutube2018}
A.~Piergiovanni and M.~S. Ryoo.
\newblock Fine-grained activity recognition in baseball videos.
\newblock In {\em CVPR Workshop on Computer Vision in Sports}, 2018.

\bibitem{qin2017zero}
J.~Qin, L.~Liu, L.~Shao, F.~Shen, B.~Ni, J.~Chen, and Y.~Wang.
\newblock Zero-shot action recognition with error-correcting output codes.
\newblock In {\em Proceedings of the IEEE Conference on Computer Vision and
  Pattern Recognition (CVPR)}, 2017.

\bibitem{rohrbach2016grounding}
A.~Rohrbach, M.~Rohrbach, R.~Hu, T.~Darrell, and B.~Schiele.
\newblock Grounding of textual phrases in images by reconstruction.
\newblock In {\em Proceedings of European Conference on Computer Vision
  (ECCV)}. Springer, 2016.

\bibitem{romera2015embarrassingly}
B.~Romera-Paredes and P.~Torr.
\newblock An embarrassingly simple approach to zero-shot learning.
\newblock In {\em International Conference on Machine Learning (ICML)}, 2015.

\bibitem{socher2013zero}
R.~Socher, M.~Ganjoo, C.~D. Manning, and A.~Ng.
\newblock Zero-shot learning through cross-modal transfer.
\newblock In {\em Advances in Neural Information Processing Systems (NIPS)},
  2013.

\bibitem{song2016unsupervised}
Y.~C. Song, I.~Naim, A.~Al~Mamun, K.~Kulkarni, P.~Singla, J.~Luo, D.~Gildea,
  and H.~A. Kautz.
\newblock Unsupervised alignment of actions in video with text descriptions.
\newblock In {\em IJCAI}, pages 2025--2031, 2016.

\bibitem{soomro2012ucf101}
K.~Soomro, A.~R. Zamir, and M.~Shah.
\newblock Ucf101: A dataset of 101 human actions classes from videos in the
  wild.
\newblock {\em arXiv preprint arXiv:1212.0402}, 2012.

\bibitem{srivastava2012multimodal}
N.~Srivastava and R.~R. Salakhutdinov.
\newblock Multimodal learning with deep boltzmann machines.
\newblock In {\em Advances in Neural Information Processing Systems (NIPS)},
  2012.

\bibitem{tran2017closer}
D.~Tran, H.~Wang, L.~Torresani, J.~Ray, Y.~LeCun, and M.~Paluri.
\newblock A closer look at spatiotemporal convolutions for action recognition.
\newblock {\em arXiv preprint arXiv:1711.11248}, 2017.

\bibitem{tygert2015convolutional}
M.~Tygert, A.~Szlam, S.~Chintala, M.~Ranzato, Y.~Tian, and W.~Zaremba.
\newblock Convolutional networks and learning invariant to homogeneous
  multiplicative scalings.
\newblock {\em arXiv preprint arXiv:1506.08230}, 2015.

\bibitem{vedantam2015cider}
R.~Vedantam, C.~Lawrence~Zitnick, and D.~Parikh.
\newblock Cider: Consensus-based image description evaluation.
\newblock In {\em Proceedings of the IEEE Conference on Computer Vision and
  Pattern Recognition (CVPR)}, 2015.

\bibitem{wang2016learning}
L.~Wang, Y.~Li, and S.~Lazebnik.
\newblock Learning deep structure-preserving image-text embeddings.
\newblock In {\em Proceedings of the IEEE Conference on Computer Vision and
  Pattern Recognition (CVPR)}, 2016.

\bibitem{wang2017zero}
W.~Wang, Y.~Pu, V.~K. Verma, K.~Fan, Y.~Zhang, C.~Chen, P.~Rai, and L.~Carin.
\newblock Zero-shot learning via class-conditioned deep generative models.
\newblock {\em arXiv preprint arXiv:1711.05820}, 2017.

\bibitem{xian2016latent}
Y.~Xian, Z.~Akata, G.~Sharma, Q.~Nguyen, M.~Hein, and B.~Schiele.
\newblock Latent embeddings for zero-shot classification.
\newblock In {\em Proceedings of the IEEE Conference on Computer Vision and
  Pattern Recognition (CVPR)}, 2016.

\bibitem{xian2018feature}
Y.~Xian, T.~Lorenz, B.~Schiele, and Z.~Akata.
\newblock Feature generating networks for zero-shot learning.
\newblock In {\em Proceedings of the IEEE Conference on Computer Vision and
  Pattern Recognition (CVPR)}, 2018.

\bibitem{xu2018joint}
H.~Xu, B.~Li, V.~Ramanishka, L.~Sigal, and K.~Saenko.
\newblock Joint event detection and description in continuous video streams.
\newblock {\em arXiv preprint arXiv:1802.10250}, 2018.

\bibitem{xu2015semantic}
X.~Xu, T.~Hospedales, and S.~Gong.
\newblock Semantic embedding space for zero-shot action recognition.
\newblock In {\em International Conference on Image Processing (ICIP)}, 2015.

\bibitem{xu2017transductive}
X.~Xu, T.~Hospedales, and S.~Gong.
\newblock Transductive zero-shot action recognition by word-vector embedding.
\newblock {\em International Journal of Computer Vision (IJCV)}, 2017.

\bibitem{yi2017dualgan}
Z.~Yi, H.~Zhang, P.~Tan, and M.~Gong.
\newblock Dualgan: Unsupervised dual learning for image-to-image translation.
\newblock In {\em Proceedings of the IEEE Conference on Computer Vision and
  Pattern Recognition (CVPR)}, 2017.

\bibitem{zhang2015zero}
Z.~Zhang and V.~Saligrama.
\newblock Zero-shot learning via semantic similarity embedding.
\newblock In {\em Proceedings of the IEEE International Conference on Computer
  Vision (ICCV)}, 2015.

\bibitem{zhou2018end}
L.~Zhou, Y.~Zhou, J.~J. Corso, R.~Socher, and C.~Xiong.
\newblock End-to-end dense video captioning with masked transformer.
\newblock {\em arXiv preprint arXiv:1804.00819}, 2018.

\bibitem{zhu2017unpaired}
J.-Y. Zhu, T.~Park, P.~Isola, and A.~A. Efros.
\newblock Unpaired image-to-image translation using cycle-consistent
  adversarial networks.
\newblock In {\em Proceedings of the IEEE Conference on Computer Vision and
  Pattern Recognition (CVPR)}, 2017.

\end{thebibliography}
\bibliographystyle{ieee}
}

\clearpage
\newpage

\appendix

\section{Implementation/training details}
We implement our models in PyTorch. For the per-segment video CNN, we use I3D \cite{carreira2017quo} to obtain a $1024\times T$ video representation. We trained a version of I3D based on Kinetics-600, but withheld all classes that appear in ActivityNet, HMDB51, or UCF101 so that the classes are truly unseen. This resulted in a training set with 478 classes and 278k videos. Since generating videos is an extremely challenging task, the video autoencoders start with and generate the I3D feature. We use GloVe word embeddings \cite{pennington2014glove} to obtain a language representation. We set $N=4$ for the temporal attention filters and apply 4 fully connected layers. These layers are followed by $L_2$ normalization so that the embedding space has unit length \cite{tygert2015convolutional}. We train the models for 200 epochs and use stochastic gradient descent with momentum to minimize the loss function with a learning rate of 0.01. After every 50 epochs, we decay the learning rate by a factor of 10. When training in the adversarial setting (e.g., Algorithm 1 in the main paper), we initialize the network training for 50 epochs on paired data followed by 200 on the paired + unpaired data.

\subsection{Unseen video captioning}

As our model learns a bi-directional mappings, we can apply our model to generate video captions. Existing video captioning models are unable to create realistic captions for unseen activities, as without training data they do not know the words to describe the video. Given a video, $v$, we can generate a caption by mapping the video to text $t=G_T(E_V(v))$. For each word, we then use nearest neighbors matching with the GloVe embeddings to obtain the words to form a sentence. In Table~\ref{res:caption}, we report the commonly used METEOR \cite{banerjee2005meteor} and CIDEr \cite{vedantam2015cider} scores of our various models, measured with the unseen classes from the ActivityNet dataset. 
We find that learning a joint representation is beneficial and using unpaired samples further improves the task. Note that this task is extremely challenging, as it requires the model to generate captions using activity words (e.g., basketball) not seen during training.

\begin{table}
  \caption{Comparison of several models for unseen activity captioning using the ActivityNet dataset, using METEOR and CIDEr scores. This evaluation was done on 10 unseen classes held out during training. Higher values are better.}
  \label{res:caption}
  \centering
  \small
  \renewcommand{\arraystretch}{0.9}
  \begin{tabular}{lcc}
    \toprule
         &  METEOR & CIDEr \\
    \midrule
    Fixed Text Representation       & 3.64  & 8.95   \\
    Joint                           & 4.21  & 9.23   \\
    All (paired)                    & 5.31  & 11.21   \\
    All (paired + unpaired)         & 6.89  & 13.95   \\
    \bottomrule
  \end{tabular}
\end{table}

\section{Additional Experiments}
\subsection{Comparison of temporal pooling methods}
To confirm that temporal attention is beneficial, we compare different forms of temporal pooling (i) max-pooling, (ii) sum-pooling, (iii) LSTM, and (iv) temporal attention filters \cite{piergiovanni2017learning}. In Table \ref{tab:temporal}, we compare these temporal pooling methods learning the joint embedding space. We confirm that using the temporal attention filters performs best.

\begin{table}
  \caption{Comparison of temporal pooling methods for 5 unseen classes in the ActivityNet dataset.}
  \label{tab:temporal}
  \centering
  \begin{tabular}{lc}
    \toprule
         &  Accuracy \\
    \midrule
    Max Pooling       & 23.4    \\
    Sum Pooling            & 24.1    \\
    LSTM              &   42.3 \\
    Temporal Attention Filters       & 55.2    \\
    \bottomrule
  \end{tabular}
\end{table}

\subsection{Comparison of different ratios of paired and unpaired data}

We compare different ratios of paired and unpaired data to see how much paired data we require and how much unpaired data is beneficial. For these experiments, we use all the loss terms (i.e., what provided us the best results). Note that in these experiments, the total number of samples was the same for each method (40k examples) so that we can directly compare the effects of unpaired data vs. paired data. Thus not all the available data was used.

In Table \ref{tab:ratio}, we show the results. We find that using no paired data results in nearly random performance, but using using some paired data greatly improves the embedding space. The model using 100\% paired data performs best, as all the others are using less overall paired data.

\begin{table}
  \caption{Comparison of different ratios of paired and unpaired data methods for 5 unseen classes in the ActivityNet dataset.}
  \label{tab:ratio}
  \centering
  \begin{tabular}{lc}
    \toprule
       Paired/Unpaired  &  Accuracy \\
    \midrule
        100\% / 0\%   &  74.2   \\
         75\% / 25\%  &  73.2    \\
         50\% / 50\%  &  69.7    \\
         25\% / 75\%  &  62.6   \\
         0\% / 100\%  &  24.5   \\
    \bottomrule
  \end{tabular}
\end{table}

We also compare augmenting our 40k paired training samples with different amounts of unpaired data. Since UCF101 and HMDB only have 13k and 7k examples, to get up to 60k samples, we also use videos from the Kinetics dataset \cite{carreira2017quo}. The results, shown in Table \ref{tab:ratio2}, show that adding the initial 10k samples is most beneficial, while additional samples do not seem to meaningfully improve results. However, due to our training method where each batch consists of 50\% paired data and 50\% unpaired data, the additional unpaired data does not harm results either.

\begin{table}
  \caption{Comparison using 40k paired examples and varying amounts of unpaired samples for 5 unseen classes in the ActivityNet dataset.}
  \label{tab:ratio2}
  \centering
  \begin{tabular}{lc}
    \toprule
       Unpaired Samples  &  Accuracy \\
    \midrule
        0  &  77.1   \\
        10k  & 82.4     \\
        20k  & 83.9     \\
        40k  & 83.6    \\
        60k  & 83.5   \\
    \bottomrule
  \end{tabular}
\end{table}

\subsection{MLB-Youtube Captions}
\begin{figure*}
    \centering
      \includegraphics[width=0.9\textwidth]{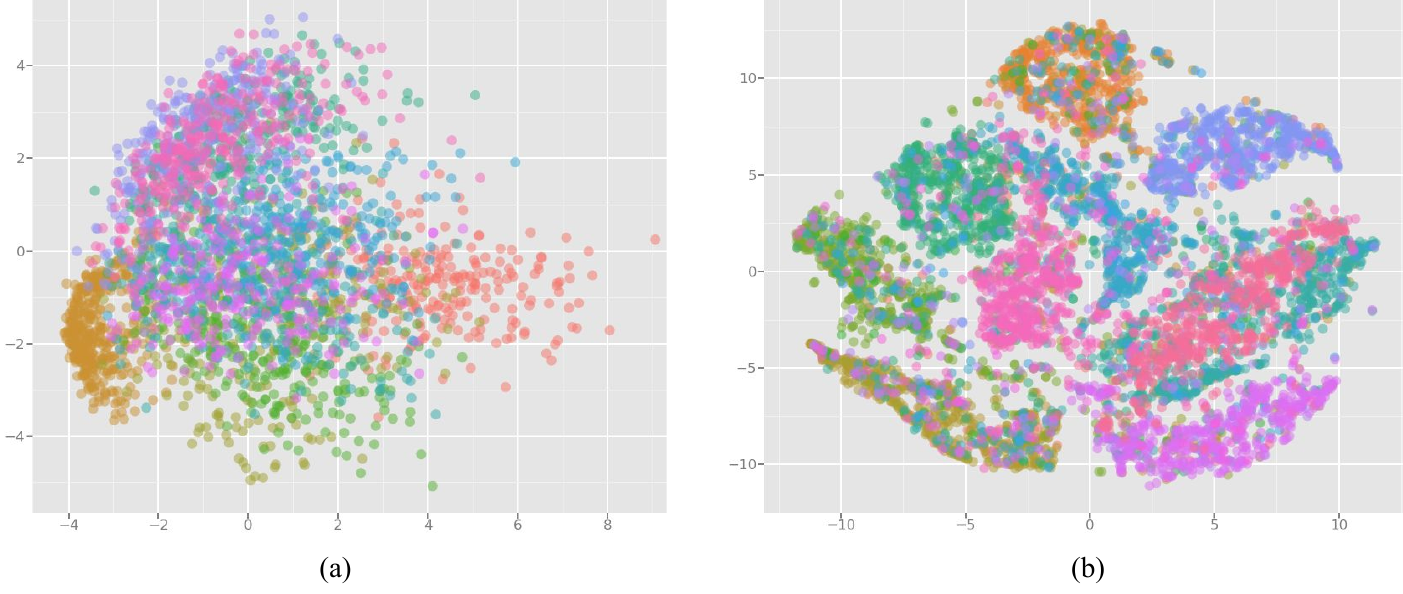}
      \caption{t-SNE mapping of \textbf{(a)} fixed text representation and \textbf{(b)} joint embedding with all paired losses for the MLB-YouTube dataset. The joint embedding space provides most distinct representations for the activities. Each color represents the activity class of the video (e.g., swing, hit, foul ball, etc.).}
      \label{fig:baseball-embed}
\end{figure*}

In Fig. \ref{fig:baseball-embed}, we compare t-SNE embeddings of the fixed text representation and our joint embedding space. This visually shows that learning a joint embedding space gives more distinct class distributions.

\subsubsection{MLB-YouTube Captions}
As a baseline for the MLB-YouTube captions dataset, we compared several different models for standard video captioning (i.e., all activity classes are seen). This task is quite challenging compared to other datasets as the announcers commentary is not always a direct description of the current events. Often the announcers tell loosely related stories and attempt to describe events differently each time to avoid repetition. Additionally, the descriptions contain on average 150 words for each 30 second interval and current captioning approaches usually only trained and tested on 10-20 word sentences. Due to these factors, this task is quite challenging the standard evaluation metrics do not account for these factors. In Table~\ref{res:caption-mlb}, we report our results on this task. 

\begin{table}
  \caption{Comparison of several models for standard, seen video captioning using the MLB-YouTube dataset, using Bleu, METEOR and CIDEr scores. Higher values are better.}
  \label{res:caption-mlb}
  \centering
  \begin{tabular}{lccc}
    \toprule
         &  Bleu & METEOR & CIDEr \\
    \midrule
    Fixed Text Representation       & 0.12 & 0.04 & 0.12   \\
    Joint Representation            & 0.14 & 0.08 & 0.15   \\
    Joint + all paired              & 0.15 & 0.10 & 0.18   \\
    Joint + paired + unpaired       & 0.10 & 0.02 & 0.08   \\
    \bottomrule
  \end{tabular}
\end{table}

\section{HMDB and UCF101 Sentences}
\label{app:sent}
For the HMBD and UCF101 datasets, we created sentences to describe each activity class. Our sentences descriptions are included in this appendix.

These sentences are written for each activity class (by randomly selecting a single video per class) and are shared for all instances of the activity. Depending on what video was randomly chosen for the class, some sentences describe the actor as a `man', `woman', or `person' which could confuse the model. Ideally, the CNN embedding needs to learn to ignore the impact of such pronoun changes.

We conducted experiments comparing randomly replacing the pronouns to determine if there was any bias introduced by the pronouns. We show the results in Table \ref{tab:pronouns}. We find that the choice of pronouns does not impact performance, as our model automatically learns to focus more on verbs rather than pronouns. When examining the temporal attention filters on the sentences, we found that they placed very little `attention' on the start of the sentence, where the pronoun usually is, suggesting that the pronoun has very little effect on the embedding space we learned.

\begin{table}
  \caption{Comparison of various pronouns on the UCF101 dataset with 50 unseen classes.}
  \label{tab:pronouns}
  \centering
  \begin{tabular}{lc}
    \toprule
         &  Accuracy \\
    \midrule
    Baseline Sentences & 33.4 \\
    All `man'       &  33.2 \\
    All `woman'       & 33.3  \\
    All `person'       & 33.4  \\
    Random pronoun       & 33.4  \\
    \bottomrule
  \end{tabular}
\end{table}

HMDB:

\begin{enumerate}
    \item chew: a woman is chewing on bread
\item golf: a man swings a golf club
\item sword exercise: a person is playing with a sword
\item walk: a person is walking
\item jump: a person jumps into the water
\item pour: a man pours from a bottle
\item laugh: a man is laughing
\item shoot gun: a person rapidly fires a gun
\item run: a person is running
\item turn: a person turns around
\item ride bike: a man is riding a bike on the street
\item swing baseball: a boy hits a baseball
\item draw sword: a person draws a sword
\item sit: a person sits in a char
\item fencing: two men are fencing
\item dribble: a boy dribbles a basketball
\item stand: a person stands up
\item pushup: a man does pushups
\item sword: two people are fighting with swords
\item pullup: a boy does pullups in a doorway
\item smile: a man smiles
\item shake hands: two people shake hands
\item shoot ball: a person shoots a basketball
\item kick: a person kicks another person
\item somersault: a person does a somersault
\item flic flac: a boy does a backflip
\item hug: two people hug
\item hit: a boy swings a baseball bat
\item dive: a person jumps into a lake
\item drink: a man drinks from a bottle
\item punch: a woman punches a man
\item wave: a person waves their hand
\item talk: a person is talking
\item kiss: a man and woman kiss
\item catch: a boy catches a ball
\item smoking: a woman smokes a cigarette
\item eat: a man eats pizza
\item throw: a person throws a ball
\item climb stairs: a man is running down the stairs
\item kick ball: a person kicks a soccer ball
\item ride horse: a girl is riding a horse
\item fall floor: a man is pushed onto the ground
\item brush hair: a girl is brushing her hair
\item situp: a man does situps
\item cartwheel: a guy runs and jumps and flips
\item pick: a man picks a book
\item push: a boy pushes a table
\item climb: a man is climbing up a wall
\item handstand: three girls do handstands
\item clap: a woman claps her hands
\item shoot bow: a person shows a bow and arrow

\end{enumerate}

UCF101:
\begin{enumerate}
    \item MilitaryParade: people are marching and waving a flag
\item TrampolineJumping: kids are jumping on a trampoline
\item PlayingDaf: a person moves a circle and hits it
\item SalsaSpin: poeple are dancing and spinning
\item CuttingInKitchen: a person is in the kitchen using a knife
\item ApplyEyeMakeup: a woman is putting on makeup
\item PlayingViolin: a person plays the violin
\item YoYo: a person plays with a yoyo
\item PlayingCello: a person is playing the cello
\item Bowling: a person is bowling
\item UnevenBars: a woman is spinning and flying on bars
\item BalanceBeam: a woman is on the balance beam
\item SkyDiving: people are falling out of the sky
\item SumoWrestling: two fat people are wrestling
\item PushUps: a man does pushups
\item FloorGymnastics: a girl does gymnastics
\item ApplyLipstick: a woman is putting on lipstick
\item BreastStroke: a woman is swimming
\item GolfSwing: a man swings a golf club
\item PlayingDohl: a person hits on a drum
\item HorseRiding: a woman rides a horse
\item PlayingFlute: a person blow into a flute
\item PizzaTossing: a man is making a pizza
\item CleanAndJerk: a person is lifting weights
\item WritingOnBoard: a person is writing on the wall
\item CricketShot: a person hits a ball with a bat
\item FieldHockeyPenalty: a girl in the field shoots a ball
\item HammerThrow: a person spins and throws an object
\item BodyWeightSquats: a man is squatting
\item CliffDiving: a person jumps off a cliff
\item Typing: a person is typing at a computer
\item MoppingFloor: a man mops the floor
\item TaiChi: people are doing tai chi
\item PlayingPiano: a person plays piano
\item Punch: someone punches another person
\item Nunchucks: a person swings nun chucks
\item RopeClimbing: a person climbs a rope
\item Swing: a baby is swinging
\item Knitting: a woman is knitting
\item Rafting: people are rafting on a river
\item PlayingGuitar: a person strums a guitar
\item ShavingBeard: a man shaves his beard
\item JugglingBalls: a person is juggling balls
\item Diving: a boy dives into a pool
\item JumpingJack: a person jumps and swings his arms
\item VolleyBallSpiking: people hit a volleyball
\item PoleValut: a person runs with a pole and launches into the air
\item SkateBoarding: a man is skateboarding
\item BoxingPunchingBag: a man is punching a bag
\item IceDancing: people are ice skating
\item WallPushups: a person does pushups against a wall
\item FrisbeeCatch: a person jumps and catches a frisbee
\item Drumming: people are drumming
\item JumpRope: a girl is jumping rope
\item HeadMassage: a person gets their head massaged
\item PlayingTabla: a person plays two drums
\item TableTennisShot: people are playing table tennis
\item PommelHorse: a person spins around on their hands
\item HighJump: a man jumps over a bar and lands on his back
\item BasketballDunk: a man jumps and dunks the basketball
\item BoxingSpeedBag: a man punches a bad in the air quickly
\item PullUps: a person does hangs on a bar and pulls up
\item RockClimbingIndoor: a person is climbing up rocks
\item BlowingCandles: a boy blows out candles on a cake
\item Skiing: people are skiing on a mountain
\item WalkingWithDog: a person walks a dog
\item Basketball: men are playing basketball
\item SoccerJuggling: a person is playing with a soccer ball
\item Fencing: people are fencing
\item Billiards: a man is playing billiards
\item BaseballPitch: a man throws a baseball
\item BlowDryHair: a woman is drying her hair
\item CricketBowling: a person throws a cricket ball
\item BandMarching: people are walking down the street playing music
\item PlayingSitar: a person plays a funny guitar
\item ThrowDiscus: a person spins and throws a disk
\item StillRings: a man holds in the air on rings
\item Lunges: a person bends to the ground with one knee
\item Skijet: a person rides a jetski in the ocean
\item BabyCrawling: a baby is crawling on the floor
\item Mixing: a woman is mixing in a bowl
\item Hammering: a person is hitting nails with a hammer
\item Shotput: a person spins and launches a ball
\item Archery: a man shoots a bow and arrow
\item Surfing: a man is surfing in the ocean
\item FrontCrawl: a person is swimming freestyle
\item HulaHoop: a person spins a hoop around their waist
\item JavelinThrow: a person throws a spear
\item Rowing: people are in a canoe and rowing
\item Kayaking: a person is kayaking on a lake
\item ParallelBars: a man does gymnastics on the parallel bars
\item HorseRace: horses are racing around a track
\item HandstandWalking: a person stands on their hands and walk
\item BrushingTeeth: a boy brushes his teeth
\item LongJump: a person runs and jumps into a sand pit
\item Biking: people are riding bikes
\item HandstandPushups: a person does pushups upside down
\item BenchPress: a man is lifting weights
\item Haircut: a person is getting a hair cut
\item TennisSwing: a woman hits a tennis ball

\end{enumerate}

\end{document}